\documentclass[journal]{IEEEtranTIE}
\usepackage{graphicx}
\usepackage{cite}
\usepackage{picinpar}
\usepackage{amsmath}
\usepackage{url}
\usepackage{flushend}
\usepackage[latin1]{inputenc}
\usepackage{colortbl}
\usepackage{soul}
\usepackage{multirow}
\usepackage{pifont}
\usepackage{color}
\usepackage{alltt}
\usepackage[hidelinks]{hyperref}
\usepackage{enumerate}
\usepackage{siunitx}
\usepackage{breakurl}
\usepackage{epstopdf}
\usepackage{pbox}
\usepackage{subfigure}
\usepackage{float}
\usepackage{booktabs}
\usepackage{threeparttable}
\usepackage{amsfonts}

\graphicspath{{Figures/}}

\begin{document}
\title{Camera-LiDAR Fusion with Latent Contact for Place Recognition in Challenging Cross-Scenes}
\author{
	Yan Pan, Jiapeng Xie, Jiajie Wu, Bo Zhou, \emph{Membership}

	\thanks{
		This work was supported by supported by the National Natural Science Foundation (NNSF) of China under the Grants No. 62073075. (Corresponding Author: Bo Zhou.)
		
		Authors are with the School of Automation, Southeast University, Nanjing 210096, P. R. China (e-mail: yanpan@seu.edu.cn; xiejiapeng@seu.edu.cn; jiajiewu@seu.edu.cn; zhoubo@seu.edu.cn). 
	}
}
\maketitle

\begin{abstract}
Although significant progress has been made, achieving place recognition in environments with perspective changes, seasonal variations, and scene transformations remains challenging. Relying solely on perception information from a single sensor is insufficient to address these issues. Recognizing the complementarity between cameras and LiDAR, multi-modal fusion methods have attracted attention. To address the information waste in existing multi-modal fusion works, this paper introduces a novel three-channel place descriptor, which consists of a cascade of image, point cloud, and fusion branches. Specifically, the fusion-based branch employs a dual-stage pipeline, leveraging the correlation between the two modalities with latent contacts, thereby facilitating information interaction and fusion. Extensive experiments on the KITTI, NCLT, USVInland, and the campus dataset demonstrate that the proposed place descriptor stands as the state-of-the-art approach, confirming its robustness and generality in challenging scenarios.
\end{abstract}



\definecolor{limegreen}{rgb}{0.2, 0.8, 0.2}
\definecolor{forestgreen}{rgb}{0.13, 0.55, 0.13}
\definecolor{greenhtml}{rgb}{0.0, 0.5, 0.0}

\section{Introduction}

\begin{figure}[t]
	\centering
	\subfigure[]{
		\centering
		\includegraphics[scale=0.2]{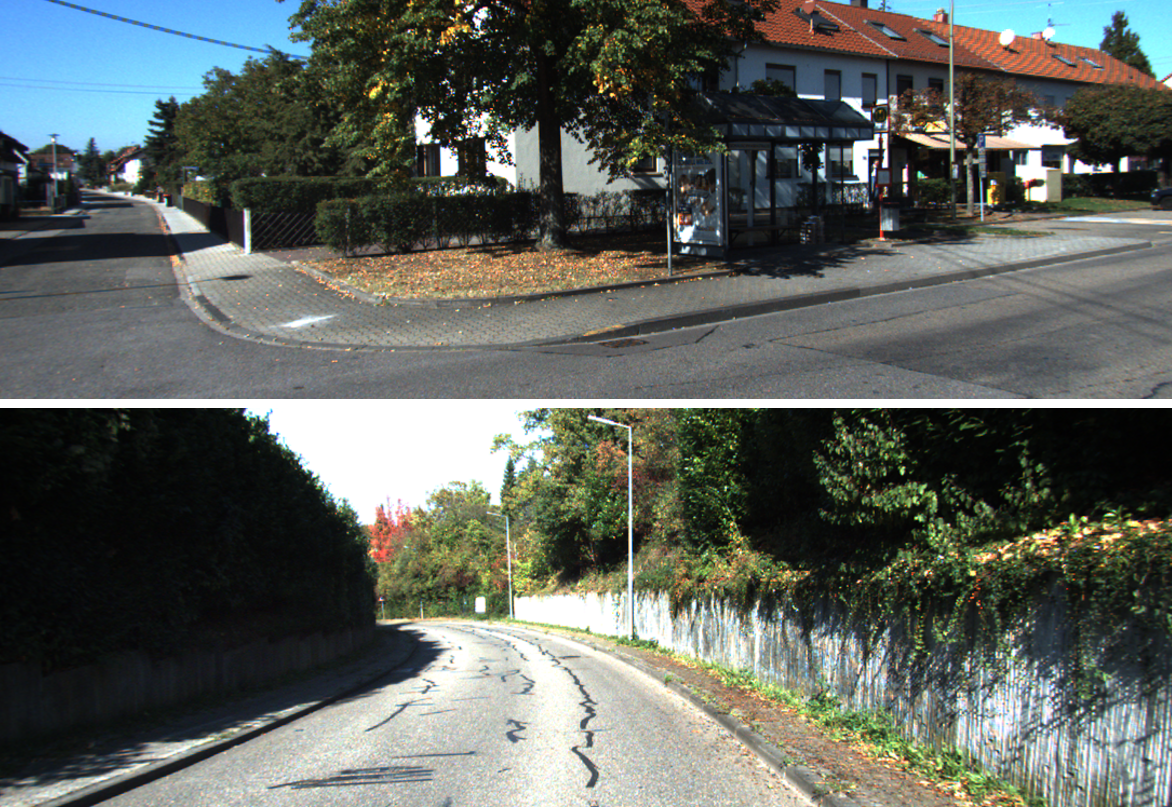}
        \label{fig:a}
	}
	\subfigure[]{
		\centering
		\includegraphics[scale=0.2]{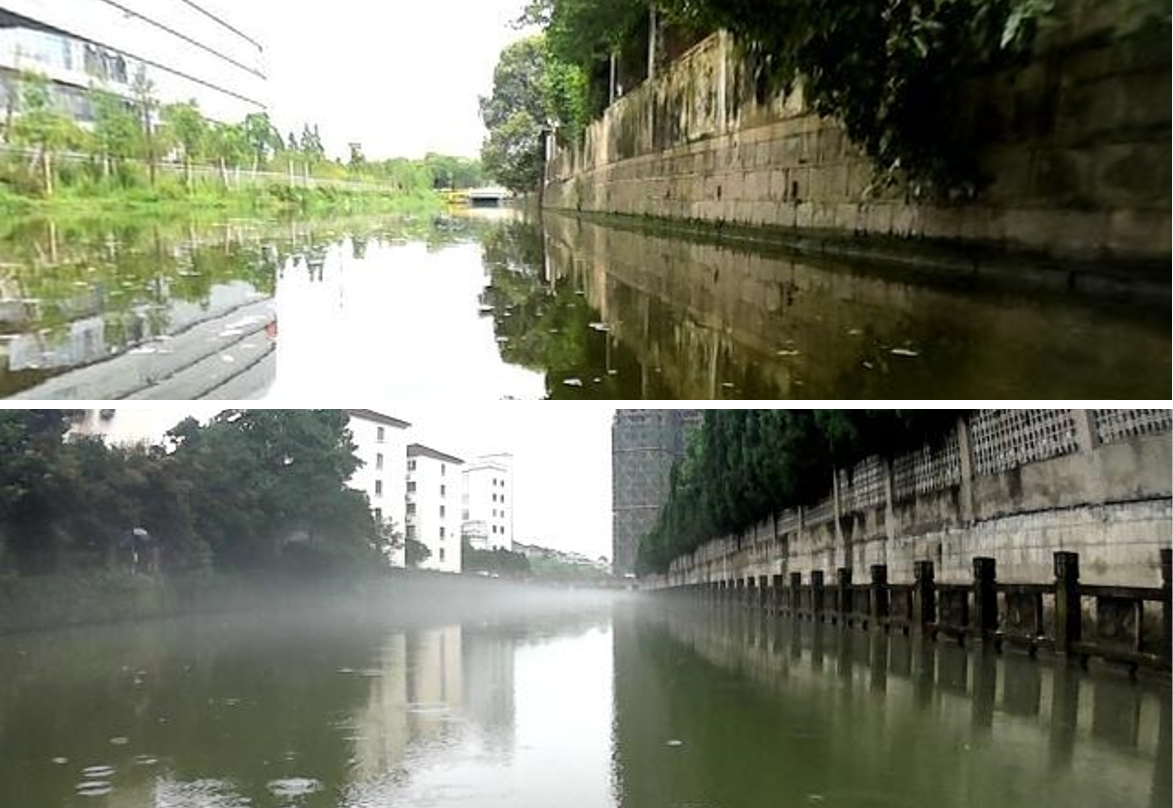}
        \label{fig:b}
	}

	\subfigure[]{
		\centering
		\includegraphics[scale=0.2]{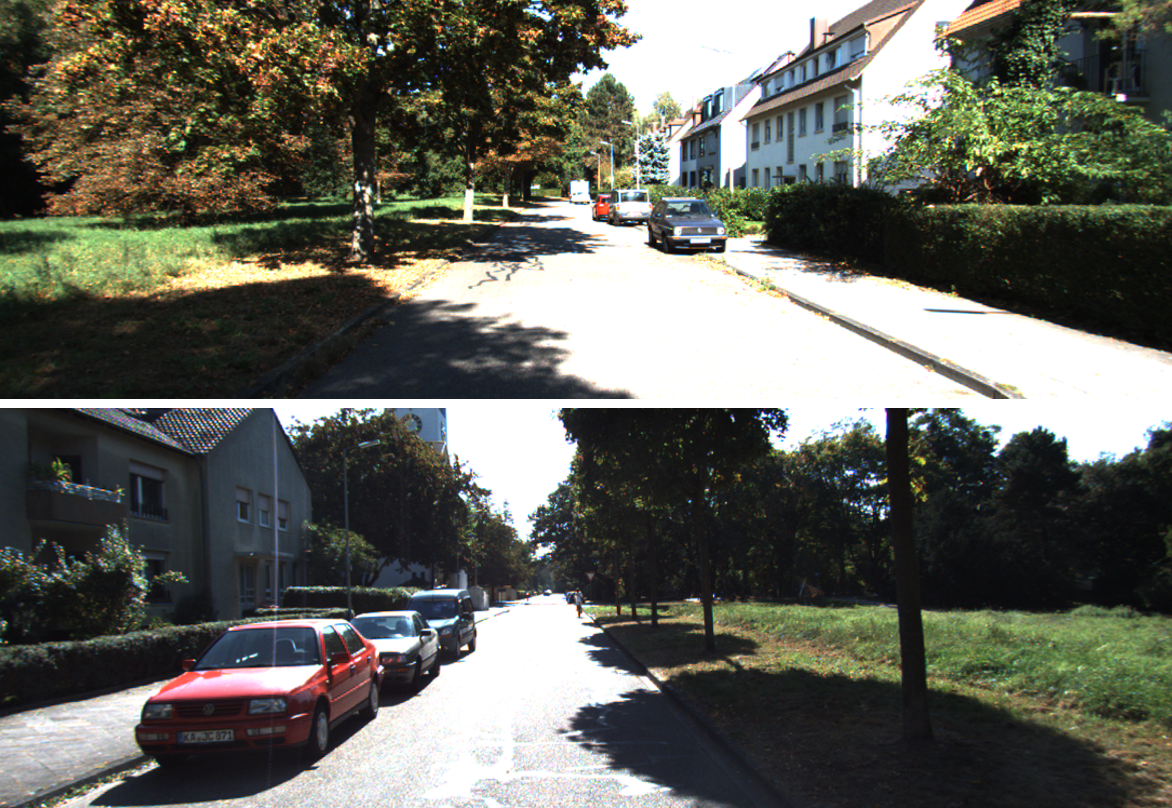}
        \label{fig:c}
	}
	\subfigure[]{
		\centering
		\includegraphics[scale=0.2]{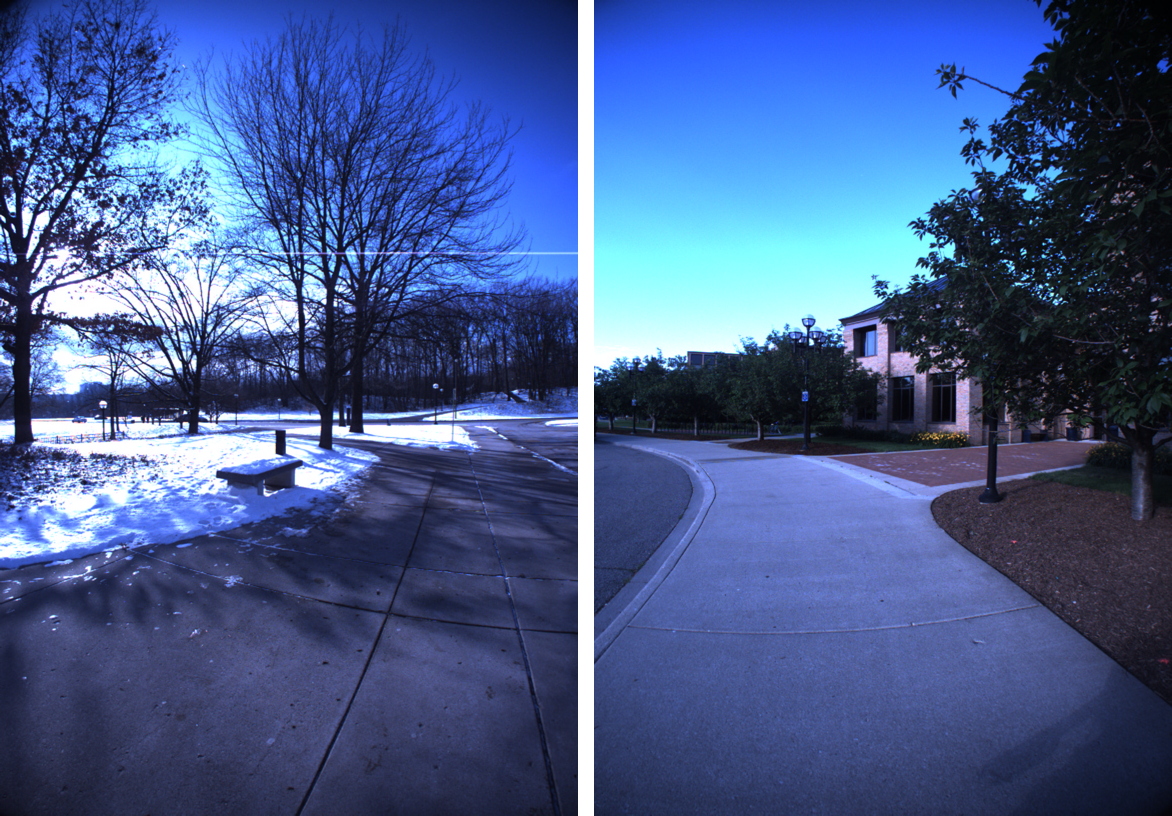}
        \label{fig:d}
	} 
    \centering
	\caption{Challenging environments in place recognition tasks: (a) an example of urban/rural roadway\cite{kitti}. (b) an example of inland waterway\cite{usvinland}. (c) an example of perspective change\cite{kitti}. (d) an example of winter/summer seasonal transition\cite{nclt}.}
    \label{challenging examples}
\end{figure}

\IEEEPARstart{P}{lace} recognition is an essential part of Simultaneous Localization and Mapping (SLAM) for robots.  Typically, it begins by characterizing the place using a descriptor, and then provides location matching by retrieving features that are most similar to the query from the database, which contributes to localization and drift error correction. However, current approaches neglect the need for cross-scene applications, and tend to be designed for a single scene, with performance dropping drastically when extended to other scenes with widely varying features. For instance, when a method designed for the urban/rural roadways as shown in Fig. \ref{fig:a} is applied to an inland waterway as shown in Fig. \ref{fig:b}, where vehicles often affected by scene characteristics such as reflections on the water surface, reflections and sparse riverbank features, and is incapable of identifying locations well. Therefore, a cross-scene place recognition method with generality is considered to be designed. In addition, the classic challenging scenarios in place recognition, such as perspective change (Fig. \ref{fig:c}) and seasonal transition (Fig. \ref{fig:d}), are also required to be addressed. For those, a place descriptor that can extract the distinctive features is crucial.

Both camera and LiDAR are two main data sources on current commonly used vehicles. Early place recognition mainly relied on rich texture information obtained from RGB cameras\cite{netvlad,patch}, but visual-based methods are inherently sensitive in the face of appearance change. In contrast, LiDAR, in the form of sparse point clouds, is able to convey geometric structural information, leading to more stable features. Several long-term localization methods based on LiDAR have emerged\cite{pointnetvlad,lpr,3dpr}, but point clouds are sparse and unordered. Evidently, the two different modalities are well complementary, and the fusion of them generally results in better performance\cite{review} to meet the challenges posed by cross-scene tasks, while there are relatively few existing approaches to combining cameras and LiDAR for place recognition.

Depending on the fusion stage, multi-modal approaches can be broadly classified into: early fusion, deep fusion and late fusion. Compared to the other two methods, deep fusion can both find the correlation between the two modal features and is relatively flexible\cite{review}. For modeling bridges between different modalities, the most common alignment approach involves obtaining correspondences between pixels and point cloud coordinates through a transformation matrix. However, due to the difference in field-of-view (FoV) of the two sensors, the explicit alignment is often limited to a forward 180-degree range, resulting in a waste of texture and structural information. To address this challenge, it is natural for us to establish a latent contact between 2D and 3D embeddings. Furthermore, while utilizing the fused features, the texture information from images and the structural information from point clouds are additionally incorporated through two separate branches.

Meanwhile, it is observed that local feature extraction has received more attention, while the equally important long-range contextual features are not easily accessible, and the learning of the latter can assist the descriptors to better represent the features. Transformer coupled with attention mechanisms\cite{transformer} provides a promising solution to this issue. Some existing efforts have introduced them into place recognition, but only to improve the extraction of visual features\cite{transvpr,hybrid} or 3D features\cite{overlaptransformer,pcan}, omitting the possibilities of further information interaction and assistance for the network to attend to important information in the scene. In our approach, the attention mechanism is employed not only to balance both local and contextual information within each modality but also to measure the relevance between the two modalities.

In sum, the contributions are as follows:

\begin{enumerate}[1)]
	\item A novel multi-modal descriptor composed of three branches: image, point cloud, and fusion. This descriptor effectively leverages the wide FoV of LiDAR and the appearance information from camera, aiming to gather diverse information supplements and enhance data utilization.
	\item A dual-stage deep fusion network to comprehensively complement and enhance information in  multi-levels. Leveraging the sequence-to-sequence cross-attention mechanism, latent correspondences between patches and points are captured, followed by a mixed-channel fusion to produce the local sub-descriptor. It is able to utilize not only the intra-dependency but also the connections across modalities.
    \item Extensive experiments on KITTI, NCLT, USVInland, and self-collected campus dataset, and comparisons with state-of-the-art single-modal and multi-modal methods, demonstrate the effectiveness and robustness of our method under perspective changes, seasonal transitions, and cross-scenes.
\end{enumerate}

\begin{figure*}
    \centering
    \includegraphics[width=\textwidth]{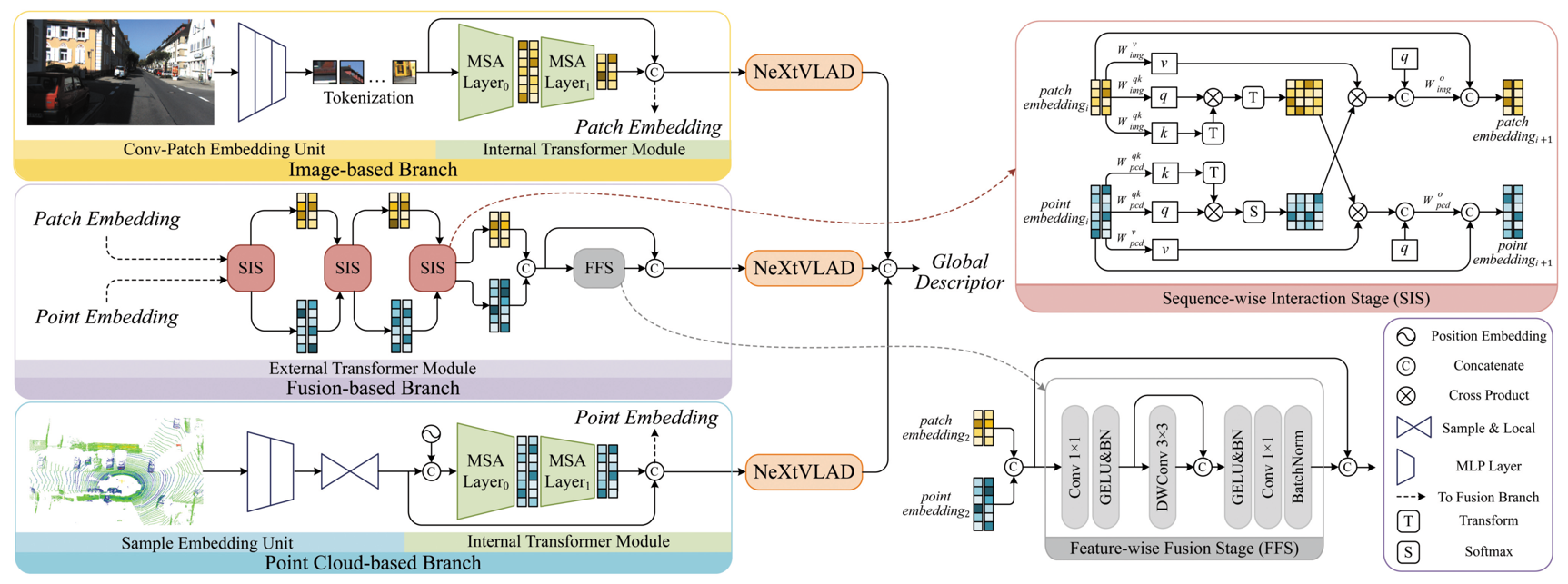}
    \caption{Overall architecture of our three-branch descriptor. It first generates local feature maps separately through the Conv-Patch Embedding Unit and Point Embedding Unit. Then, the corresponding Internal Transformer Module encodes features into two local sub-descriptors. Subsequently, the dual-stage External Transformer Module in the fusion-based branch is employed to establish the latent contact. Initially, the Sequence-wise Interaction Stage learns the cross-modal correlations, followed by the Feature-wise Fusion Stage, which combines the stacked features through channel mixing. Finally, NeXtVLAD is used on three branches respectively to generate the three-branch global descriptor for place recognition.}
    \label{fig:pipeline}
\end{figure*}

\section{Related Works}
\subsection{Visual-based Place Recognition (VPR)}
Early classical VPR methods primarily relied on techniques, such as BoW\cite{bow} and VLAD\cite{vlad}, to aggregate local handcrafted features. However, thanks to the advancements in deep learning, learning-based place recognition algorithms have shown superior performance. A widely adopted method is NetVLAD\cite{netvlad}, which has inspired various variants. It initially extracts features through a pre-trained CNN network, then employs VLAD\cite{vlad} to aggregate the fragment features into a global vector. The limitation lies in constructing a global descriptor solely from the single representation output, resulting in relatively lower recall rates. Considering the incorporation of multi-scale information, Patch-NetVLAD\cite{patch} performs patch-level image matching based on NetVLAD, enhancing the ability to handle severe appearance changes.

Global modeling is critical in environments where perspectives change dramatically, while CNN-based methods are limited to extracting dense local features. Fortunately, Transformer\cite{transformer} provides a flexible and adaptable structure to tackle this issue. Some work has extended Transformer based on CNN to capture context-aware features\cite{cvt}. TransVPR\cite{transvpr} merge features to form patch-level descriptors, and identify task-relevant features at different semantic levels with multi-head self-attention. Multi-scale features from CNNs and hierarchical information from Transformers are integrated into Hybrid CNN-Transformer\cite{hybrid}, selecting informative cues of global characteristics with semantic priors. 


\subsection{LiDAR-based Place Recognition (LPR)}
Compared with the camera, LiDAR benefits from capturing the geometric structure of the surroundings, which is robust against variable conditions. Scan Context\cite{scancontext} encodes point clouds from bird's eye view, facilitating faster and more efficient context search. Although some fine-grained depth information is sacrificed, it retains height information which proves valuable for long-term localization.

Building upon the development in 3D object detection, researchers combined PointNet\cite{pointnet} and NetVLAD\cite{netvlad} to create PointNetVLAD\cite{pointnetvlad}, which successfully transposed VPR into the LPR. LPD-net\cite{lpdnet}, on the other hand, employs graph-based aggregation modules in both coordinate and feature spaces to extract local features from point clouds by combining handcrafted features.

Inspired by VPR, the attention mechanism is introduced to predict the significance of features. For instance, PCAN\cite{pcan} employs an attention map for better concentration on crucial features, leading to discriminative global descriptors. OverlapTransformer\cite{overlaptransformer} and its subsequent sequence-enhanced method SeqOT\cite{seqot} adopt overlap instead of distances, and establish a yaw-angle-invariant descriptor that exhibits superior performance in scenarios with orientation changes.

\subsection{Fusion-based Place Recognition (FPR)}
Recently, the complementarity between camera and LiDAR has been recognized, prompting researchers to emphasize the importance of the fusion of two distinct modalities. MinkLoc++\cite{minkloc++} adopts a late fusion approach by individually processing each modality before integrating them in the final stage. MMDF\cite{mmdf}, the deep fusion method, capitalizes on the spatial relationships between pixels and point clouds to fuse the two modalities.

As the popularity of Transformers has grown, attention mechanisms have been introduced into multi-modal place recognition. AdaFusion\cite{adafusion} utilizes self-attention modules to generate adaptive weights, which dynamically adjust the contributions of both modalities in different environments. PIC-Net\cite{picnet} employs the global channel attention to enhance the representational capabilities of image and point cloud features, ultimately producing the final global descriptor. 

Different from the above methods, the attention mechanism in our approach is applied in both intra-modality and inter-modality relationships. Within a single modality for images or point clouds, it is utilized to capture contextual information to help extract more representative features. Between the two modalities, it instead focuses on discovering potential relationships between image regions and point cloud features, thereby capturing latent cross-modal cues between these two different modalities, taking into account both texture features and structural features.

\section{Our Approach}
Our network architecture is illustrated in Fig.\ref{fig:pipeline}. The place descriptor consists of the image-based branch, point-based cloud branch, and fusion-based branch. In Section III-A, the input raw data from camera and LiDAR are tokenized to obtain Patch Embedding and Point Embedding. Then, in Section III-B, the \textit{Internal Transformer Module} is introduced to model intra-modality relationships, resulting in single-modal local sub-descriptors for images and point clouds. Section III-C presents the \textit{External Transformer Module}, which enhances the ability of the fusion-based branch to obtain distinctive sub-descriptors through \textit{Sequence-wise Interaction Stage} and \textit{Feature-wise Fusion Stage}. The aggregation of these three local sub-descriptors and concatenation into a global descriptor is explained in Section III-D. Finally, Section III-E provides a brief overview of the loss functions and metric learning.

\subsection{Input Data Preparation}
\subsubsection{Conv-Patch Embedding Unit} 
Traditional Transformer-based visual methods often split the image into patches directly\cite{vit}, and this tokenization process can lead to the neglect of details found in edges and corners. To address this limitation, some methods consider gathering patch embeddings from feature maps\cite{incorporating}. Building upon this idea, we build the Conv-Patch Embedding Unit for the initial encoding of image regions, which allows us to utilize the advantages of CNN in capturing correlating neighboring pixels. Given the input image $I\in\mathbb{R}^{H\times W\times3}$, where $H$ and $W$ are the height and width respectively, three conservative $3 \times 3$ convolutions are used, where the first convolution has a stride of 2. The resulting local feature map $I\in\mathbb{R}^{\frac{H}{2} \times \frac{W}{2}\times C}$, where $C$ represents the number of channels, and the value is set as 16 in all experiments. This approach reduces the number of parameters and computational costs while ensuring efficient local feature extraction, even for images with lower resolutions.

For processing the local feature map, we follow the successful practice from vision Transformers\cite{vit}. Concretely, each image is evenly partitioned into a sequence of patches  $p\in\mathbb{R}^{P\times P\times C}$, and the length of the sequence is $N=\frac{H}{P} \times \frac{W}{P}$, which is also the number of patches, where P represents the patch size. In contrast to those methods used for classification or segmentation, we remove the position embedding since spatial information is already implicitly encoded within the patch embedding.

\subsubsection{Point Cloud Embedding Unit}
Similar to the image branch, the local feature map is first acquired, upon which tokenization is applied. With the aim of preserving the intrinsic geometric features of raw point clouds, the Point Cloud Embedding Unit is designed following the ideas of PointNet++\cite{pointnet++} and PCT\cite{pct}. When given the original point cloud $P\in\mathbb{R}^{N_0\times3}$ as input, two cascaded MLP layers which consist of a $1\times1$ convolution layer followed by a BatchNorm layer and a ReLU layer are initially employed. 

Assuming the input point cloud feature as $f_p$, the farthest point sampling algorithm is employed to sample points as the center for each local neighborhood, with coordinates $p_s$ and features $f_s$. Instead of the query ball\cite{pointnet++}, the k-Nearest Neighbors is utilized to expand the sampled points into sampling neighborhoods $\mathcal{N}_i\in\mathbb{R}^{k\times\left(3+D\right)}$, where D is the dimension of $f_p$, by considering their $k$ nearest neighboring points with coordinates $p_q$ and features $f_q$. This approach can effectively avoid the imbalanced point distribution within neighborhoods caused by the irregularity of input point clouds. 

Finally, the features and coordinates of output points, denoted as $p_o$ and $f_o$ respectively, can be computed as follows:
\begin{equation}
    \delta_{f_{qs}}=f_q-f_s\ ,\ \delta_{p_{qs}}=p_q-p_s
\end{equation}
\begin{equation}
    h_\Theta\left(\delta_{f_{qs}},\delta_{p_{qs}}\right)=h_\Theta[\delta_{f_{qs}};\delta_{p_{qs}}]
\end{equation}
\begin{equation}
    f_o=MaxPool\left(h_\Theta\left(\delta_{f_{qs}},\delta_{p_{qs}}\right)\right),\ p_o=p_s
\end{equation}
where $\delta_{f_{qs}}$ and $\delta_{p_{qs}}$ represent the differences between neighboring points and sampled points in the coordinate space and feature space, respectively, and $[\cdot;\cdot]$ denotes the concatenation operation. $h_\Theta(\cdot)$, the feature encoding function comprises two cascaded MLP layers, and serves to bring semantically similar points closer together in the embedding space. Here, MLP layers share the same parameters with the former one. 

Specifically, the number of the point embedding is decreased to 512 and 256, while the feature dimension is increased to 128 and 256, respectively. Because the aggregated coordinates of the points encompass the real-world spatial relationships, we directly apply them as positional encodings.

\subsection{Internal Transformer Module}
The acquisition of distinguishable feature representations plays a crucial role in scene understanding. To establish rich contextual relationships among local features, patch embeddings and point embeddings are fed into the Internal Transformer Module.

Comprising the self-attention mechanism, this module can be described as a mapping of the input query and a collection of key-value pairs. It calculates an output by weighting the values using contacts between the query and its associated key. Drawing inspiration from the Transformer\cite{transformer}, Multi-head Self-Attention(MSA) with $h$ indexes over $H=4$ heads is employed for gathering information from various representational spaces. Given the stacked patch embeddings or point embeddings $x$, we first calculate the query, key and value for the input by dot product with learnable linear projections $W_h^q$, $W_h^k$, $W_h^v$. Subsequently, the dot product between the query and key is computed, followed by a softmax layer to normalize the attention score $\alpha^h$: 
\begin{equation}
    \alpha^h=softmax\frac{\left(W_h^qx\right)\left(W_h^kx\right)^T}{\sqrt{d_k}}
\end{equation}
where $\alpha^h$ represents the correlation between the query and key, and is regarded as the weight of value. Furthermore, the weighted sum of value for all heads is computed and concatenated, followed by a projection $W_o$ to the output:
\begin{align}
    SelfAttn\left(x\right)=W^o [\alpha^1 \cdot W_1^v v; \ldots; \alpha^H \cdot W_H^v v]
\end{align}

\subsection{External Transformer Module}
After obtaining the local sub-descriptor of images and point clouds, we build a dual-stage External Transformer Module to enhance the information interaction and merge the features from two modalities into the sub-descriptor of fusion-based branch. As shown in Fig. \ref{fig:pipeline}, in Sequence-wise Interaction Stage, image regions and point clouds are contextually attended to each other, achieving information exchange based on their similarity. In Feature-wise fusion stage, the stacked features are further merged through a channel embedding.

\subsubsection{Sequence-wise Interaction Stage}
In this stage, relevant information is interacted with based on the latent connections between image regions and point cloud features. When performing information interaction between two modalities, one modality is projected as query while the other modality is mapped as key and value. Since the modalities can have different feature sequence lengths, the transformation matrix $W_h^{qk}$ and $W_h^v$ is used to align the dimensions. Also, there are $h=4$ parallel heads are employed in our method. Taking the image $I$ as the query and the point cloud $P$ as the key and value, for instance, the vectors are formed as follow:
\begin{align}
    Q_h=W_h^{qk}I,\ K_h=W_h^{qk}P,\ V_h=W_h^vP
\end{align}

Next, we perform matrix multiplication between $Q_h$ and the transpose of $K_h$, and apply a softmax layer to normalize. This generated a cross-attention matrix, which models the latent contact between features of two modalities:
\begin{align}
    s_{PI}^h=softmax\frac{Q_hK_h^T}{\sqrt{d_k}}
\end{align}
where $s_{PI}^h$ implies the correlation between point cloud features and image regions. Through a weighted sum and a projection $W^o$ to the output dimension, the interacted image features are calculated as:
\begin{equation}
    CrossAttn\left(I\right)=W^o \sum _{h=1}^H (s_{PI}^h)
\end{equation}
This selective aggregation of similar features effectively enhances the semantic consistency between the two modalities. Subsequently, the attended image features and the residual one are concatenated and passed through a residual Feed-Forward Neural Network layer.

\subsubsection{Feature-wise Fusion Stage}
In the second stage, fusing two types of information at the channel level is considered rather than a simple concatenation. Additionally, it is noted that a significant amount of information is lost during the convolution process, which can result in a decrease in the ability of the sub-descriptor to describe the place. Motivated by Mobilenetv2\cite{mobilenetv2}, we introduce a depth-wise convolution layer, DW Conv $3\times3$, to achieve an inverted residual structure with negligible extra computational cost, and modify the location of the shortcut connection.

\subsection{Metric Learning}
In the field of place recognition, metric learning techniques are employed due to the absence of specific classes. Here we use the common triplet loss function\cite{triplet} to train the multi-modal place descriptors. Assuming a triplet tuple$ \mathcal{T}=\left(P_{query},P_{pos},P_{neg}\right)$, with $P_{query}$, $P_{pos}$, $P_{neg}$ denoting the global features of the query, positive, and negative places, respectively. Specifically, the positive place represents a true match, meaning that its ground truth is within a distance of $d_{pos}$ from the query, and the negative place denotes the false match, whose Euclidean distance to the query is beyond $d_{neg}$. In all of our experiments, we set $d_{pos}=5m$ and $d_{neg}=50m$, and the number of positive and negative samples is ensured to be equal in the training and test sets.

For each triplet tuple, the triplet loss function aims to minimize the distance between the global descriptor vectors of $P_{query}$ and $P_{pos}$, while maximizing the distance between the global descriptor vectors of $P_{query}$ and $P_{neg}$:
\begin{align}
    \mathcal{L}_{triplet}\left(\mathcal{T}\right)=\left[margin+\delta_{pos}-\delta_{neg}\right]_+
\end{align}
where $[x]_+=max\left(x,0\right)$ denotes the hinge loss, and $\delta_{pos}=d(P_{query},P_{pos})$ while $\delta_{neg}=d(P_{query},P_{neg})$), ensuring that semantically similar samples are closer to the query place than any different negative samples.

\section{Experiments}
\subsection{Datasets}
We evaluated the performance of our proposed framework on KITTI\cite{kitti}, NCLT\cite{nclt}, USVInland\cite{usvinland}, and a self-collected campus dataset named SEU-s. Details regarding the dataset are presented in the Table \ref{tab:dataset}.

\begin{table}[htbp]
  \centering
  \caption{sequences of four datasets used in experiments}
    \begin{tabular}{ccc}
    \toprule
    Dataset & Sequences & Conditions \\
    \midrule
    \multirow{5}[2]{*}{KITTI} & 00    & Urban / Same-direction \\
          & 02    & \textbf{Mainly rural} / Mainly Same-direction  \\
          & 05    & Urban / Same-direction / Intersection \\
          & 06    & Urban / Same-direction \\
          & 08    & Urban / \textbf{Reverse-direction} \\
    \midrule
    \multirow{6}[2]{*}{NCLT} & 2012-01-08 & Midday / Partly Cloudy \\
          & 2012-01-15 & Afternoon / Sunny / \textbf{Snow} \\
          & 2012-02-05 & Morning / Sunny \\
          & 2012-06-15 & Morning / Sunny / \textbf{Foliage} \\
          & 2012-12-01 & Evening / Sunny \\
          & 2013-02-23 & Afternoon / Cloudy / \textbf{Snow} \\
    \midrule
    \multirow{2}[2]{*}{USVInland} & H05\_9\_115\_700 & \textbf{Mist} \\
          & N03\_2\_80\_536 & Overcast \\
    \midrule
    \multirow{3}[2]{*}{SEU-s} & 00    & Campus / - \\
          & 01    & Campus / Same-direction \\
          & 02    & Campus / \textbf{Reverse-direction} \\
    \bottomrule
    \end{tabular}%
  \label{tab:dataset}%
\end{table}%

\textbf{KITTI} This dataset provides high-resolution RGB images and LiDAR point clouds in the city, rural areas, and highways. Sequences 00, 02, 05, 06, and 08 which contain loop closures are selected. The first pass served as a database, while the second pass along the same paths was used as a query. To increase the number of tuples in the training set, sequences 07 and 09, which do not contain loop closures, are included.

\textbf{NCLT} This dataset covers diverse outdoor conditions involving weather, season and illumination changes, and provides point clouds from LiDAR and images from six individual cameras. In our experiments, images from the front-facing Cam 5 are employed. Using the timestamps of LiDAR, the associated images and coordinates with the closest timestamp are found. Our method, along with all baseline methods, is trained on sequences 2012-01-08 and 2012-01-05. These two sequences are used as the database, while the four new sequences, 2012-02-05, 2012-06-15, 2012-12-01, and 2013-02-23, are used as queries.

\textbf{USVInland} This is a dataset for inland waterways collected under various weather conditions, comprising data from stereo cameras and LiDAR. We selected two sequences from the public dataset, namely N03\_2\_80\_536(Overcast) and H05\_9\_115\_700(Mist), both of which have loop closures. Similar to KITTI, the first pass through the revisit is regarded as the database and the second pass is used as the query.

\begin{figure}[htbp]
    \centering
    \includegraphics[width=0.65\linewidth]{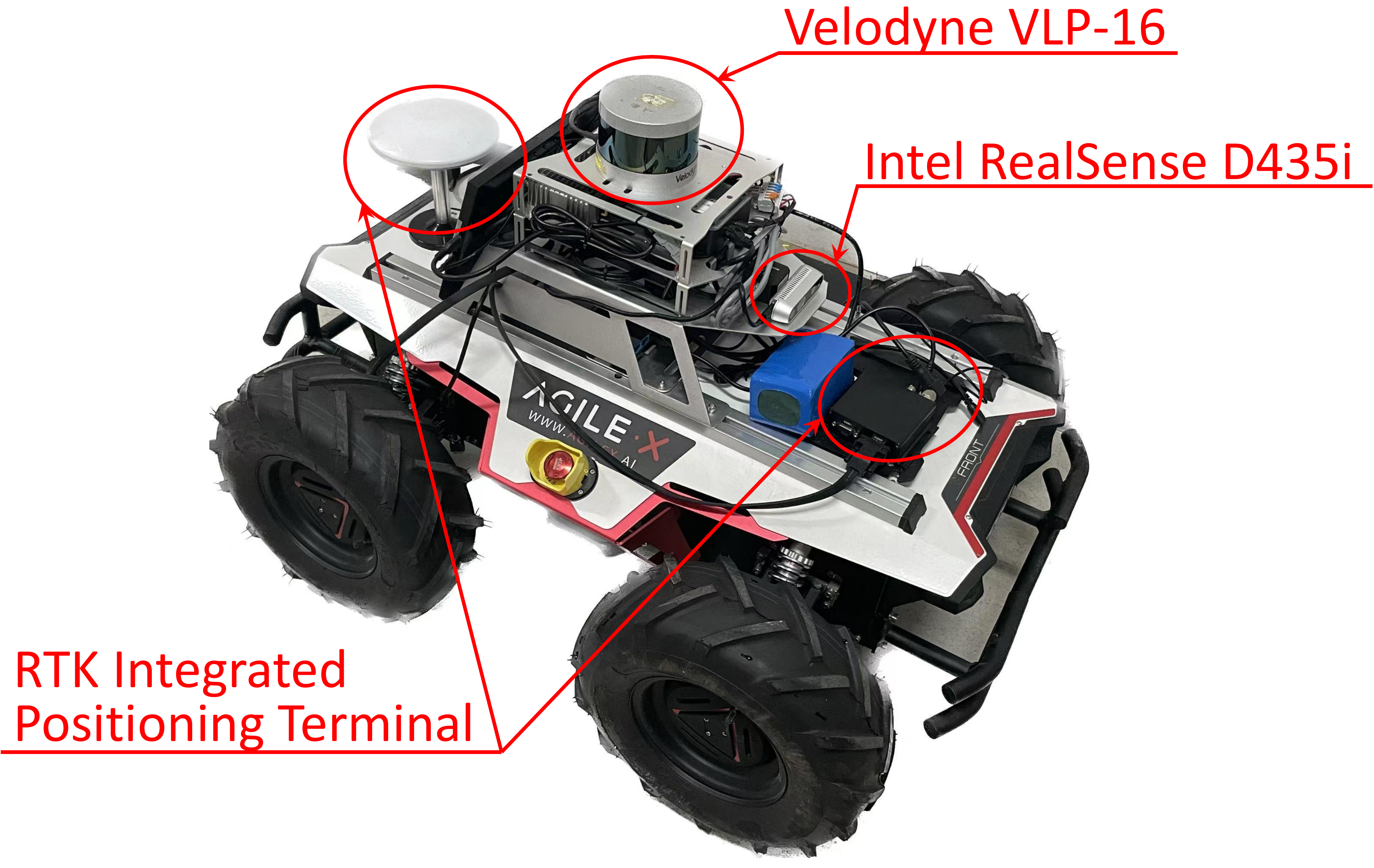}
    \caption{The unmanned vehicle used for data collection. }
    \label{fig:ugv}
\end{figure}

\textbf{SEU-s}  Based on an Intel RealSense D435i visual sensor, a Velodyne VLP-16 LiDAR sensor, and the RTK integrated positioning terminal, we collected data within our campus environment utilizing the unmanned vehicle as Fig.\ref{fig:ugv}. Inspired by NCLT, data collection was performed along the same route but in different driving directions. Sequence 00 is regarded as a database, and sequence 01 and 02 are queries.

\subsection{Implementation Details}
The proposed Network is implemented in PyTorch framework with a NVIDIA RTX 3090 GPU. The whole network is trained with the batch size of 16, the SGD optimizer with a momentum of 0.5, a learning rate of $10^{-4}$, and a weight decay of $5\times10^{-4}$. The model that exhibited the best performance after 100 epochs was selected. In the preprocessing of input data, we scaled images to a resolution of $320\times96$, and point clouds are downsampled to 8192 points. For all multi-head attention mechanisms within the method, the number of heads IS set to 4, and dropout is configured as 0. Regarding Global Descriptor Generator, the width multiplier $\lambda$ is set to 2, and the size of groups is 8. With 64 clusters, we finally obtain three 256-dimensional sub-descriptors after aggregations. 

Various place recognition methods are compared with ours, including VPR: NetVLAD (NV)\cite{netvlad}, LPR: PointNetVLAD (PNV)\cite{pointnetvlad}, OverlapTransformer (OT)\cite{overlaptransformer}, and FPR: MinkLoc++\cite{minkloc++}, MMDF\cite{mmdf}. For a fair comparison, all these methods are evaluated by ourselves through available code or pre-trained models. Precision-Recall curve, a well-recognized evaluation metric for place recognition in the community, is used to evaluate the performances, while the maximum F1 score is used to assess the Precision-Recall curves further.

\subsection{Place Recognition Under Viewpoint Change}
\begin{figure}[htbp]
    \centering
    \includegraphics[width=0.8\linewidth]{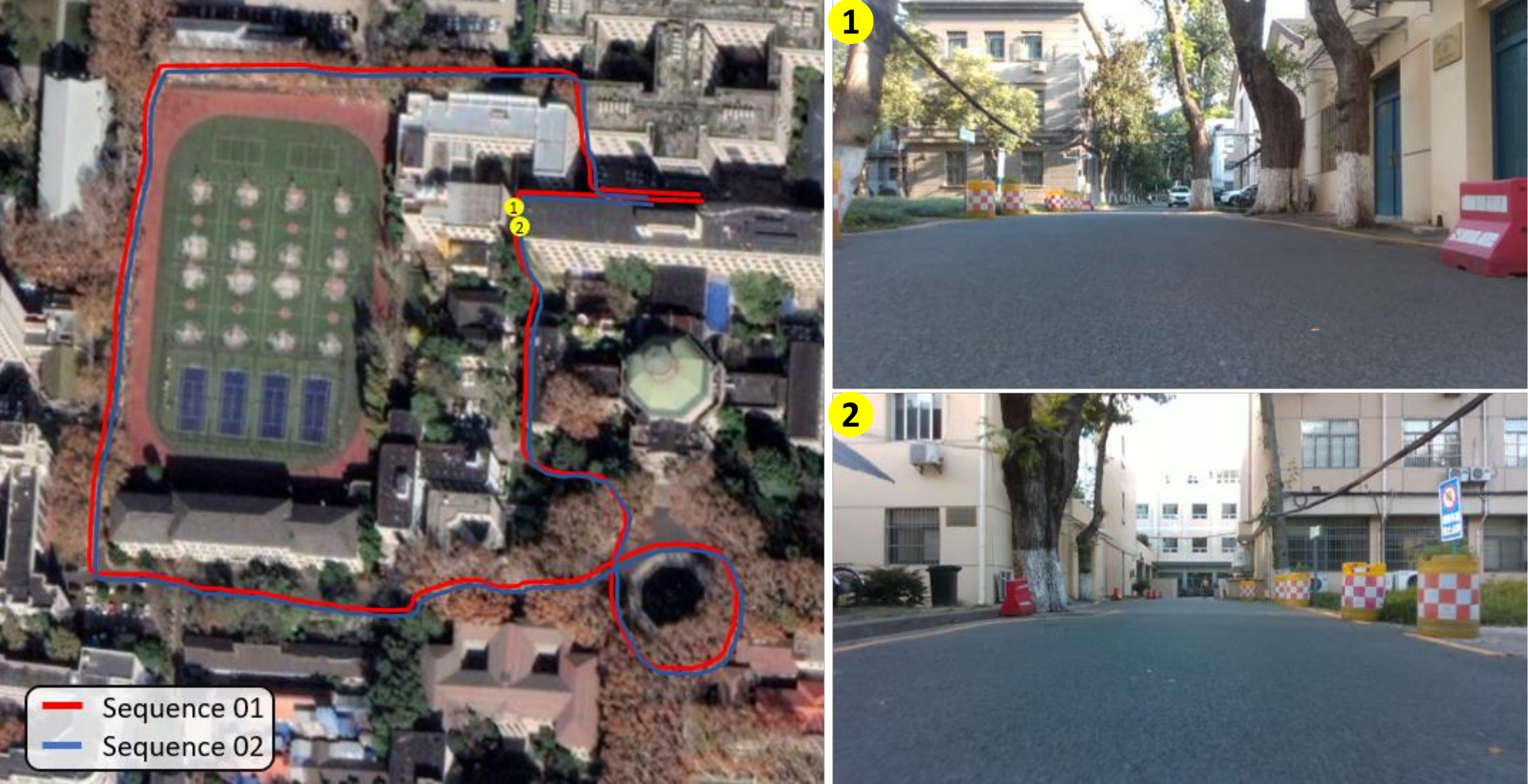}
    \caption{Data collection for campus. Left: Trajectories for sequences 01 and 02. Right: Images taken while passing through the same place from the opposite direction.}
    \label{fig:campus}
\end{figure}

In this subsection, our method and all the baseline methods are trained only on the KITTI dataset, and directly evaluated on both the KITTI and SEU-s datasets, demonstrating the effectiveness and making comparisons with baseline methods under viewpoint changes. From the KITTI dataset, we selected typical paths including same-direction revisits, opposite-direction revisits, and intersections. As shown in Fig. \ref{fig:campus}, sequence 01 from SEU-s dataset is the same-direction revisit and sequence 02 is in the reverse direction.

\begin{figure}[htbp]
	\centering
	\includegraphics[width=0.95\linewidth]{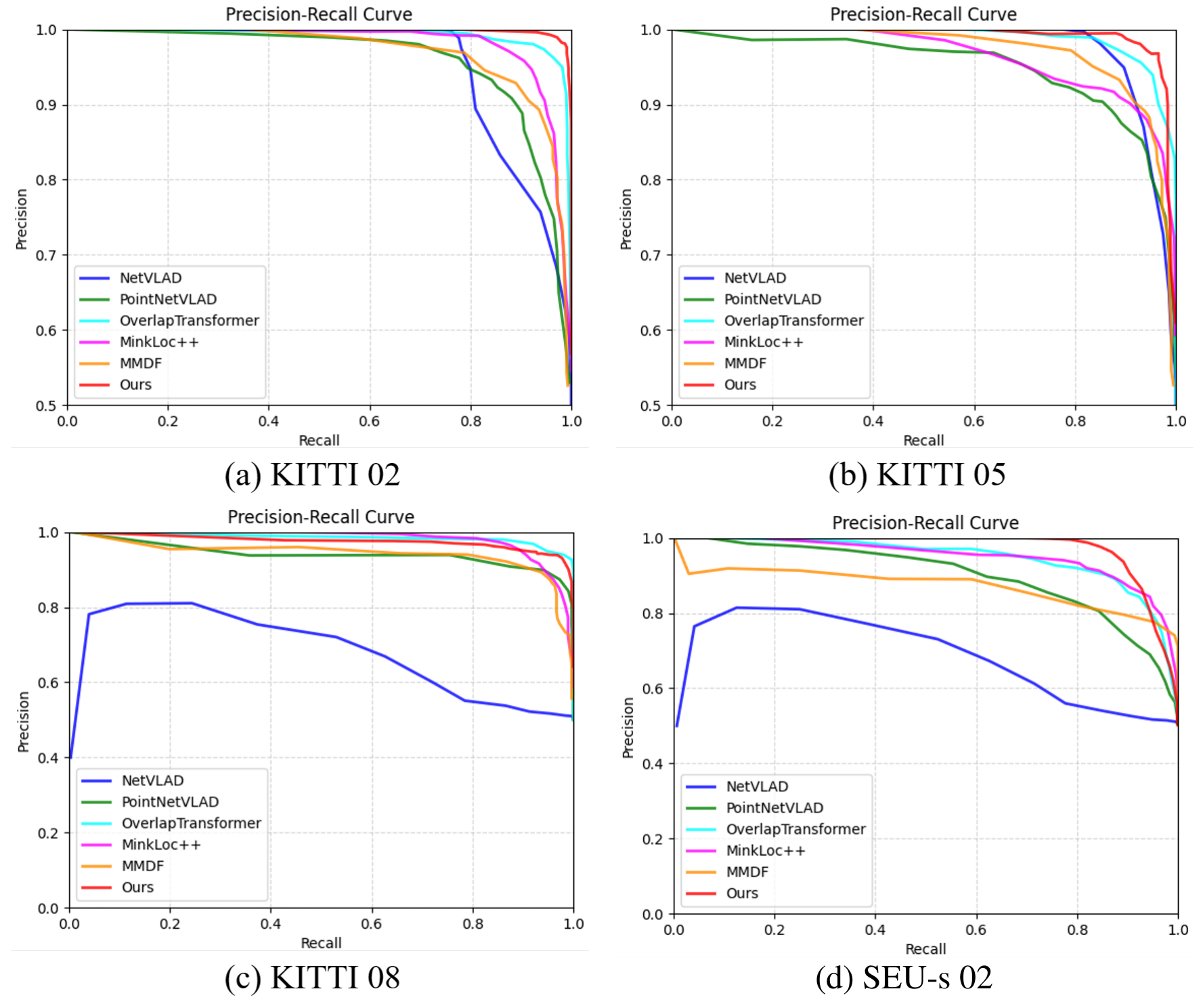}
	\caption{Comparison of Precision-Recall curves on KITTI and SEU-s.}
	\label{fig:kitti}
\end{figure}

\begin{table}[htbp]
  \centering
  \setlength{\tabcolsep}{1.8pt}
  \caption{comparison of maximum F1 scores on KITTI and SEU-s dataset}
  \begin{threeparttable}
    \begin{tabular}{ccccccccc}
    \toprule
    \multirow{2}[4]{*}{Methods} & \multicolumn{1}{c}{\multirow{2}[4]{*}{Modality}} & \multicolumn{5}{c}{KITTI}             & \multicolumn{2}{c}{SEU-s} \\
\cmidrule{3-9}          &       & 00     & 02     & 05     & 06     & 08     & 01    & 02 \\
    \midrule
    NetVLAD    & V     & 0.955 & 0.871 & 0.923 & 0.971 & 0.676 & 0.897 & 0.675 \\
    PointNetVLAD   & L     & 0.909 & 0.829 & 0.786 & 0.924 & 0.921 & 0.842 & 0.820 \\
    OverlapTransformer    & L     & 0.964 & 0.966 & 0.947 & 0.979 & \textbf{0.962} & 0.918 & 0.884 \\
    MinkLoc++ & F     & 0.946 & 0.934 & 0.912 & 0.961 & 0.924 & 0.921 & 0.892 \\
    MMDF  & F     & 0.948 & 0.915 & 0.937 & 0.965 & 0.914 & 0.927 & 0.856 \\
    Ours  & F     & \textbf{0.973} & \textbf{0.984} & \textbf{0.967} & \textbf{0.996} & 0.954 & \textbf{0.939} & \textbf{0.897} \\
    \bottomrule
    \end{tabular}%
    \begin{tablenotes}
        \footnotesize
        \item V: Visual, L: LiDAR, F: Fusion of Visual and LiDAR.
    \end{tablenotes}
    \end{threeparttable}
  \label{tab:kitti}%
\end{table}%

Table \ref{tab:kitti} presents the maximum F1 scores on both KITTI and campus datasets, while Fig. \ref{fig:kitti} selectively shows precision-recall curves for KITTI 02, 05, 08, and SEU-s 02, where KITTI 08 and SEU-s 02 are reverse revisits. It can be observed that on KITTI 00, 02, 05, 06, and SEU-s 01, which are primarily same-direction revisits, our method exhibits better stability and precision than others. The 06 sequence, in particular, involves a long-distance forward revisit, and single-modal and multi-modal fusion methods already perform well on it. Building upon this, our method still exceeds the best-performing method by 1.7\%, achieving a maximum F1 score of 0.996. In contrast, sequence 02 represents a scenario with sparse features that pose a greater challenge for place recognition, and NV, PNV, and MMDF show a decrease compared to the other three sequences. However, our multi-modal fusion method even outperforms with a recall rate of up to 79\% when the precision rate is 100\%.

In the case of KITTI 08 and SEU-s 02, the latter exhibits slightly decreased performance compared to the former overall. This is attributed to the fact that there are many similar locations in SEU-s, making it easier to confuse the perception of scenes. Moreover, it is evident from Fig. \ref{fig:kitti}(c)(d) that image-only methods perform poorly, while point cloud-only and multi-modal methods perform relatively better on two opposite-direction revisits. This is because the FoV of the camera is limited, and image-only methods are susceptible to viewpoint variations. As for multi-modal methods, deep fusion-based MMDF focuses its feature extraction on overlapped areas to establish a connection between two modalities, whereas MinkLoc++ performs late fusion at the strategy stage and is less influenced by the image. Notably, OT achieves a slightly higher maximum F1 score than other point-based methods, and this can be attributed to the range image used as input in OT, providing more comprehensive and complete information. To ensure computational efficiency, point-based methods, such as PNV, MinkLoc++, MMDF, and our method, employ downsampling, and the negative impact in performance is noticeable when the number of points decreases\cite{minkloc3d}.

\subsection{Place Recognition Across Season}
In this subsection, we choose to validate our methods on the NCLT dataset in the face of seasonal variations. Rich vegetation in the summer and snow in the winter make the change in appearance more dramatic during seasonal variations. Therefore, we select sequences with a time span longer than a year, with a focus on summer and winter sequences. Moreover, NCLT is collected at different times, and the heading is randomly selected, they inherently introduce challenges related to variations in lighting conditions and changes in perspective.

\begin{figure}[htbp]
    \centering
    \includegraphics[width=0.95\linewidth]{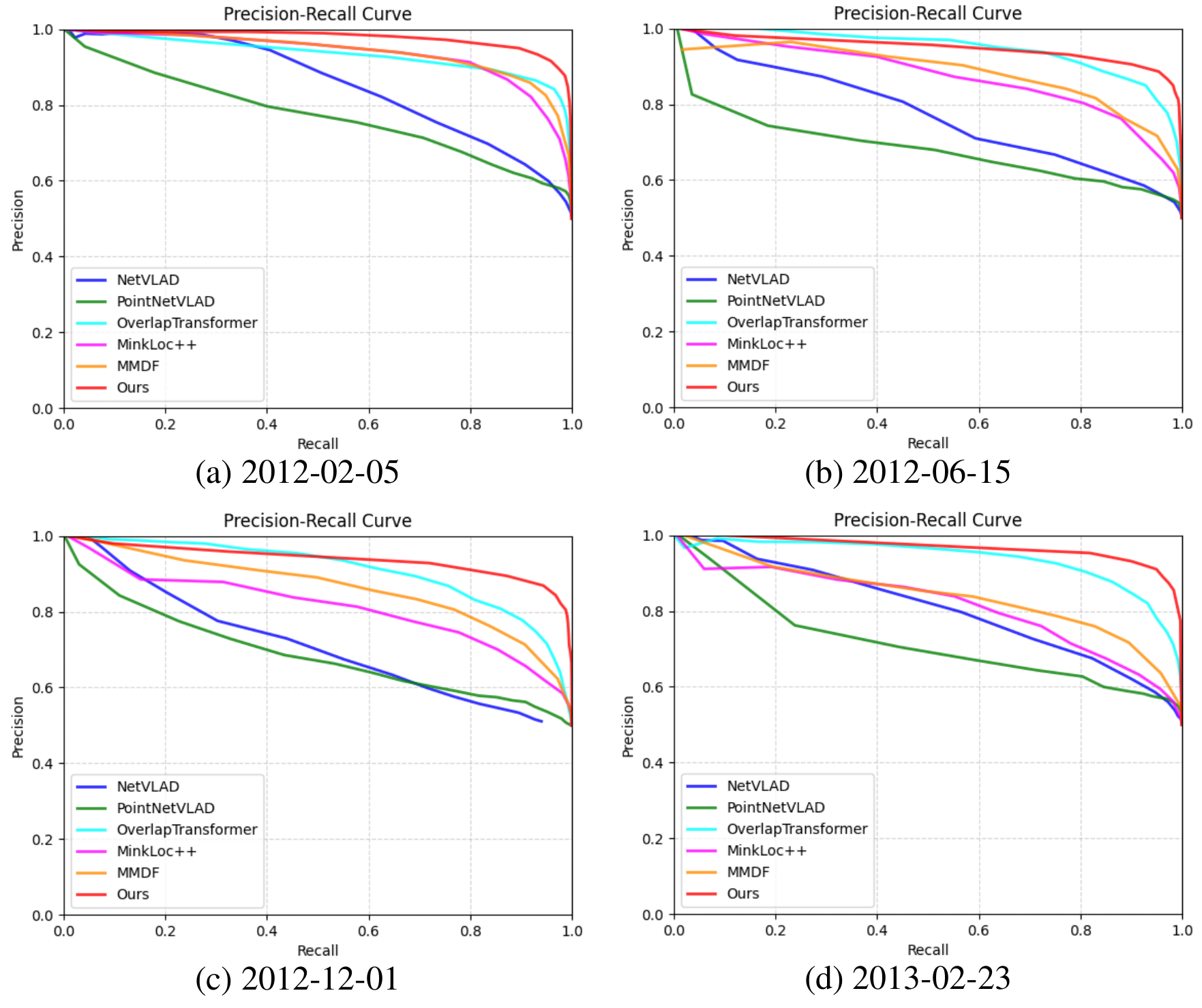}
    \caption{Comparison of Precision-Recall curves on NCLT.}
    \label{fig:nclt}
\end{figure}

\begin{table}[htbp]
  \centering
  \setlength{\tabcolsep}{2pt}
  \caption{comparison of maximum F1 scores on NCLT dataset}
  \begin{threeparttable}
    \begin{tabular}{ccccccc}
    \toprule
    Methods & Modality  & \multicolumn{1}{p{2.735em}}{2012-\newline{}02-05} & \multicolumn{1}{p{2.735em}}{2012-\newline{}06-15} & \multicolumn{1}{p{2.735em}}{2012-\newline{}12-01} & \multicolumn{1}{p{2.735em}}{2013-\newline{}02-23} & Avg. \\
    \midrule
    NetVLAD & V     & 0.760 & 0.721 & 0.669 & 0.742 & 0.723 \\
    PointNetVLAD & L     & 0.731 & 0.708 & 0.694 & 0.717 & 0.713 \\
    OverlapTransformer & L     & 0.899 & 0.887 & 0.840 & 0.873 & 0.875 \\
    MinkLoc++ & F     & 0.872 & 0.817 & 0.769 & 0.754 & 0.803 \\
    MMDF  & F     & 0.887 & 0.823 & 0.800 & 0.798 & 0.827 \\
    Ours  & F     & \textbf{0.937} & \textbf{0.919} & \textbf{0.904} & \textbf{0.929} & \textbf{0.922} \\
    \bottomrule
    \end{tabular}%
    \begin{tablenotes}
        \footnotesize
        \item V: Visual, L: LiDAR, F: Fusion of Visual and LiDAR.
    \end{tablenotes}
    \end{threeparttable} 
  \label{tab:nclt}%
\end{table}%

As shown in Fig. \ref{fig:nclt} and Table \ref{tab:nclt}, our method performs relatively worse on NCLT compared to the KITTI and SEU-s datasets. This can be explained by the fact, that because training and test sets of KITTI and SEU-s are in the similar condition, and yet the NCLT training set lacks foliage and does not perfectly match the weather conditions of the test set. As the time interval between training and test sequences increases, significant changes occur in the features of each sequence. However, our method continues to exhibit superior precision-recall performance in challenging perception environments, with an average maximum F1 score improvement of $4.7\%$ compared to the best-performing methods. Furthermore, it is evident that when dealing with summer sequence (2012-06-15) and winter sequence (2012-12-01), other methods experience significant declines in performance, while our method is less affected, with performance fluctuations of less than $3.3\%$. This result confirms that our method achieves excellent generalization and robustness in place recognition even across seasons.

\subsection{Place Recognition in cross-scenes}
In the USVInland dataset, we intend to validate the effectiveness of our method on water surfaces, demonstrating its versatility in cross-scenes. H05\_9\_115\_700 in a mist day and N03\_2\_80\_536 in an overcast day are selected, and we add the recall of each method when the precision is $100\%$ for exploring the potential for practical applications.



\begin{figure}[htbp]
    \centering
    \includegraphics[width=0.95\linewidth]{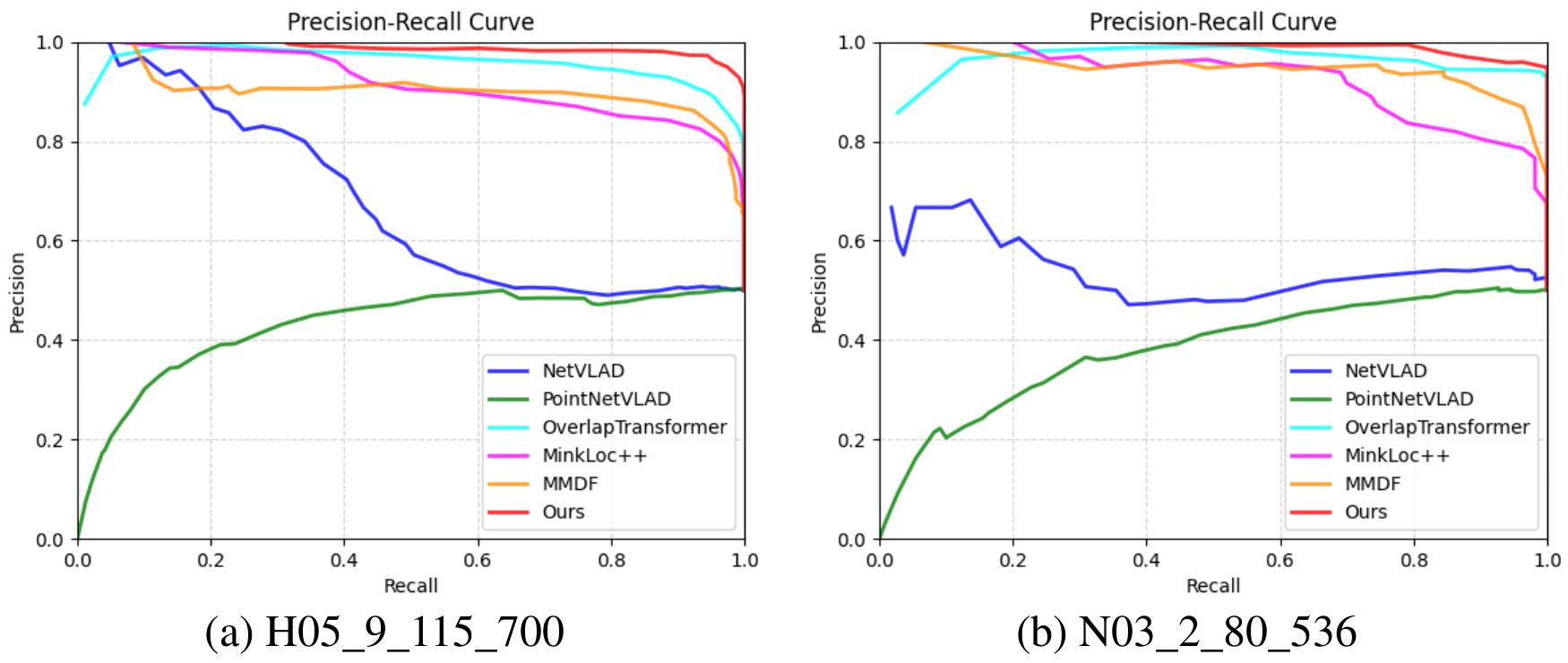}
    \caption{Comparison of Precision-Recall curves on USVInland.}
    \label{fig:usvinland}
\end{figure}

\begin{figure}[htbp]
    \centering
    \includegraphics[width=0.95\linewidth]{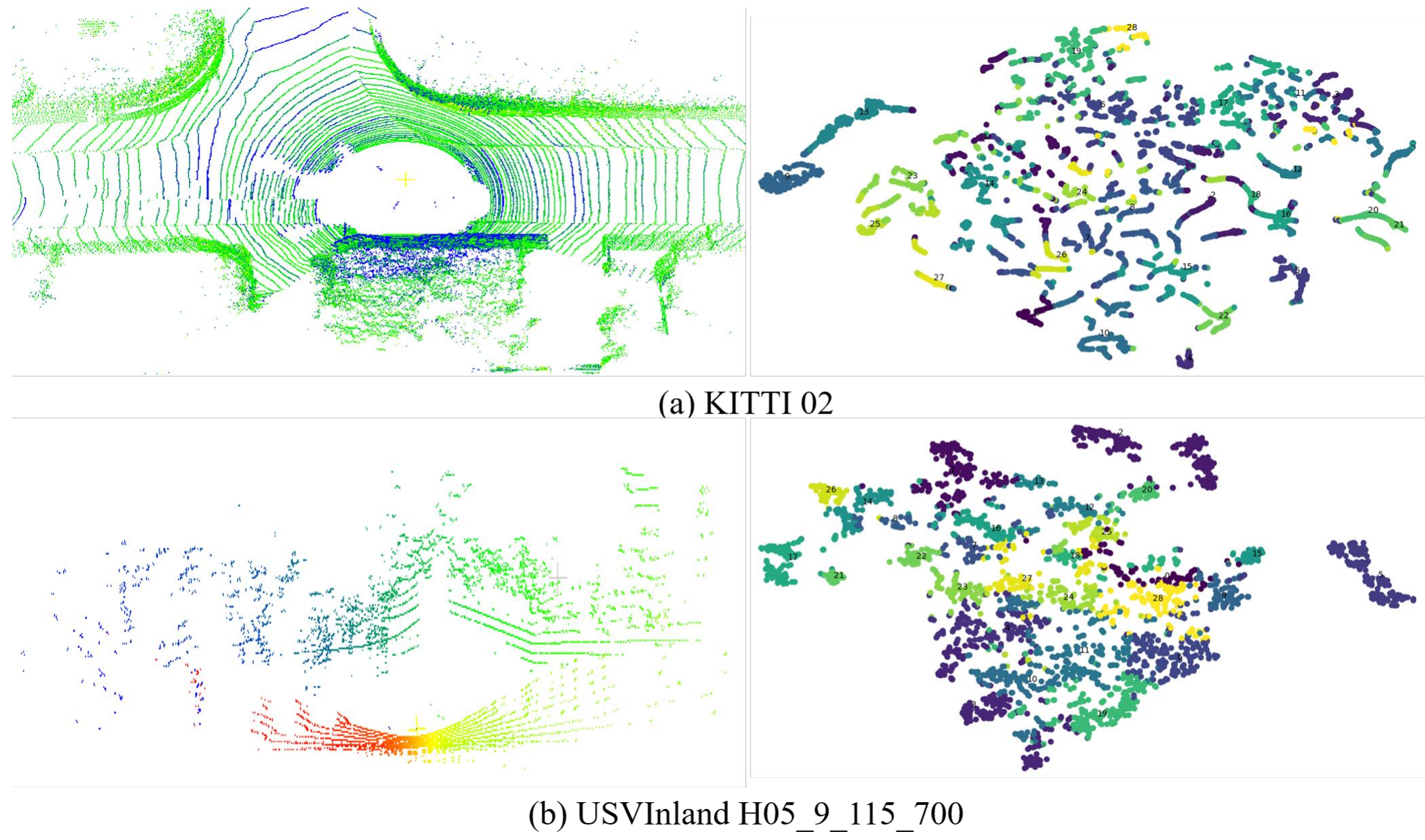}
    \caption{Point clouds and visualization of place clustering using t-SNE on our descriptors for KITTI 02 and USVInland H05\_9\_115\_700, respectively. The point clouds on the water surface are much sparser than those on the ground, while the inter-class difference is smaller than that on the ground, indicating that contextual features in waterways are more similar.}
    \label{fig:place-cluster}
\end{figure}

As shown in Fig. \ref{fig:usvinland} and Table \ref{tab:usvinland}, classic single-modal algorithms like NV and PNV, commonly used in autonomous vehicles, experience a significant drop in performance. As shown in Fig.\ref{fig:place-cluster}, This drop is attributed to the absence of suitable matching features in inland waterways where the surroundings are often dense vegetation and slopes. Additionally, LiDAR is sensitive to rain and fog due to its operating principles, and the water reflections also impact visual features mapping. Consequently, these single-modal methods lack the capability to generate distinguishable descriptors of inland waterways.

On the other hand, MinkLoc++ and MMDF, which fuse images and point clouds, demonstrate the significant improvement in place recognition capabilities offered by multi-modal approaches. In particular, MMDF leverages deep fusion to establish a tighter connection between two modalities and achieves a better performance. Besides, when precision is set at $100\%$, the recall of ours is $30.9\%$ and $37.7\%$, respectively, while the recall of OT drops significantly, approaching zero.

\begin{table}[htbp]
  \centering
  \setlength{\tabcolsep}{4pt}
  \caption{max recall rates when precision rates is $100\%$ of place recognition methods on USVInland dataset}
  \begin{threeparttable}
    \begin{tabular}{cccc}
    \toprule
    Methods & Modality & H05\_9\_115\_700 & N03\_2\_80\_536 \\
    \midrule
    NetVLAD & V     & 4.7\% & N/A \\
    PointNetVLAD & L     & N/A   & N/A \\
    OverlapTransformer & L     & N/A   & N/A \\
    MinkLoc++ & F     & 6.2\% & 20.7\% \\
    MMDF  & F     & 8.2\% & 20.1\% \\
    Ours  & F     & \textbf{30.9\%} & \textbf{37.7\%} \\
    \bottomrule
    \end{tabular}%
    \begin{tablenotes}
        \footnotesize
        \item V: Visual, L: LiDAR, F: Fusion of Visual and LiDAR.
    \end{tablenotes}
    \end{threeparttable} 
  \label{tab:usvinland}%
\end{table}%

\subsection{Ablation Study}
In this subsection, we first investigate the effects of different branches and their combinations on place recognition tasks through ablation experiments. Table \ref{tab:abl1} presents the maximum F1 scores for image-only, point cloud-only, fusion-only, concatenation of images and point clouds, and our method on KITTI 00 and NCLT 2012-06-15. Obviously, the introduction of multi-modalities, whether through simple concatenation or advanced fusion method, indeed makes the descriptor more identifiable, especially in challenging environments. Moreover, the three-branch approach helps compensate for some of the lost features due to the implicit alignment of modalities, further improving the robustness.

\begin{table}[htbp]
  \centering
  \caption{Ablation study on the effects of different branches and their combinations on KITTI and NCLT dataset}
    \begin{tabular}{ccc}
    \toprule
    Structures & KITTI 00 & NCLT  2012-06-15\\
    \midrule
    Visual Branch & 0.928 & 0.734 \\
    Lidar Branch & 0.936 & 0.747 \\
    Fusion Branch & 0.961 & 0.892 \\
    Visual+Lidar (concat) & 0.943 & 0.873 \\
    Visual+Lidar+Fusion (Ours) & \textbf{0.973} & \textbf{0.919} \\
    \bottomrule
    \end{tabular}%
  \label{tab:abl1}%
\end{table}%

In the External Transformer Module, two stages are applied to interact and combine features from images and point clouds. To explore the effectiveness of SIS and FFS, we conducted experiments on the KITTI 00 sequence for these two stages separately, and the results are shown in Table \ref{tab:abl2}. If SIS is removed, it means directly using the features independently extracted in their respective branches. As for FFS, the simple concatenation is employed to replace feature fusion. Compared to the baseline, there is a $1.9\%$ improvement in SIS-only approach, and a $0.5\%$ increase using only FFS. If both SIS and FFS are contained, the F1 max scores will be improved by $3.0\%$. In other words, utilizing cross-attention mechanisms for inter-modal information exchange in SIS yields a greater improvement compared to using FFS.

\begin{table}[htbp]
  \centering
  \caption{ablation study on SIS and FFS on KITTI sequence 00}
    \begin{tabular}{ccc}
    \toprule
    Feature Interaction & Feature Fusion & F1 max scores \\
    \midrule
    No    & Concat & 0.943 \\
    SIS   & Concat & 0.964 \\
    No    & FFS   & 0.948 \\
    SIS   & FFS   & \textbf{0.973} \\
    \bottomrule
    \end{tabular}%
  \label{tab:abl2}%
\end{table}%

\section{Conclusion}
This paper introduces a novel multi-branch cascaded network that combines data from LiDAR and cameras to enhance place recognition capabilities in challenging environments. First, our three-branch global descriptor effectively leverages all information from the camera and LiDAR with different FoV, thus improving information utilization. Second, we creatively establish a latent cross-modal contact, enabling fine-grained information interaction based on the correlation between different modalities, followed by channel-level fusion to generate the fusion-based sub-descriptor. Extensive experiments on the KITTI, NCLT, USVInland, and SEU-s datasets demonstrate the superior performance of our method compared to state-of-the-art single-modal and multi-modal place recognition methods. Our approach stands out in precision-recall curves and maximum F1 score metrics, validating its feasibility in challenge scenarios involving viewpoint changes, seasonal transitions, and cross-scenes. In future work, we consider introducing semantic information as a constraint and researching human-like place recognition methods.

\bibliographystyle{Bibliography/IEEEtranTIE}
\bibliography{IEEEabrv,Bibliography/reference}\ 

\begin{thebibliography}{10}
\providecommand{\url}[1]{#1}
\csname url@samestyle\endcsname
\providecommand{\newblock}{\relax}
\providecommand{\bibinfo}[2]{#2}
\providecommand{\BIBentrySTDinterwordspacing}{\spaceskip=0pt\relax}
\providecommand{\BIBentryALTinterwordstretchfactor}{4}
\providecommand{\BIBentryALTinterwordspacing}{\spaceskip=\fontdimen2\font plus
\BIBentryALTinterwordstretchfactor\fontdimen3\font minus
  \fontdimen4\font\relax}
\providecommand{\BIBforeignlanguage}[2]{{%
\expandafter\ifx\csname l@#1\endcsname\relax
\typeout{** WARNING: IEEEtran.bst: No hyphenation pattern has been}%
\typeout{** loaded for the language `#1'. Using the pattern for}%
\typeout{** the default language instead.}%
\else
\language=\csname l@#1\endcsname
\fi
#2}}
\providecommand{\BIBdecl}{\relax}
\BIBdecl

\bibitem{kitti}
A.~Geiger, P.~Lenz, and R.~Urtasun, ``Are we ready for autonomous driving? the
  kitti vision benchmark suite,'' in \emph{Proc. IEEE Comput. Soc. Conf.
  Comput. Vision. Pattern. Recognit.}, pp. 3354--3361.\hskip 1em plus 0.5em
  minus 0.4em\relax IEEE, 2012.

\bibitem{usvinland}
Y.~Cheng, M.~Jiang, J.~Zhu, and Y.~Liu, ``Are we ready for unmanned surface
  vehicles in inland waterways? the usvinland multisensor dataset and
  benchmark,'' \emph{IEEE Robot. Autom. Lett.}, vol.~6, no.~2, pp. 3964--3970,
  2021.

\bibitem{nclt}
N.~Carlevaris-Bianco, A.~K. Ushani, and R.~M. Eustice, ``University of michigan
  north campus long-term vision and lidar dataset,'' \emph{Int. J. Rob. Res.},
  vol.~35, no.~9, pp. 1023--1035, 2016.

\bibitem{netvlad}
R.~Arandjelovic, P.~Gronat, A.~Torii, T.~Pajdla, and J.~Sivic, ``Netvlad: Cnn
  architecture for weakly supervised place recognition,'' in \emph{IEEE Trans.
  Pattern. Anal. Mach. Intell.}, pp. 5297--5307, 2016.

\bibitem{patch}
S.~Hausler, S.~Garg, M.~Xu, M.~Milford, and T.~Fischer, ``Patch-netvlad:
  Multi-scale fusion of locally-global descriptors for place recognition,'' in
  \emph{Proc. IEEE Comput. Soc. Conf. Comput. Vision. Pattern. Recognit.}, pp.
  14\,141--14\,152, 2021.

\bibitem{pointnetvlad}
M.~A. Uy and G.~H. Lee, ``Pointnetvlad: Deep point cloud based retrieval for
  large-scale place recognition,'' in \emph{PProc. IEEE Comput. Soc. Conf.
  Comput. Vision. Pattern. Recognit.}, pp. 4470--4479, 2018.

\bibitem{lpr}
D.~Kong, X.~Li, Y.~Hu, Q.~Xu, A.~Wang, and W.~Hu, ``Learning a novel lidar
  submap-based observation model for global positioning in long-term changing
  environments,'' \emph{IEEE Trans. Ind. Electron.}, vol.~70, no.~3, pp.
  3147--3157, 2023.

\bibitem{3dpr}
P.~Yin, F.~Wang, A.~Egorov, J.~Hou, Z.~Jia, and J.~Han, ``Fast
  sequence-matching enhanced viewpoint-invariant 3-d place recognition,''
  \emph{IEEE Trans. Ind. Electron.}, vol.~69, no.~2, pp. 2127--2135, 2022.

\bibitem{review}
Y.~Peng, Y.~Qin, X.~Tang, Z.~Zhang, and L.~Deng, ``Survey on image and
  point-cloud fusion-based object detection in autonomous vehicles,''
  \emph{IEEE Trans. Intell. Transport. Syst.}, vol.~23, no.~12, pp.
  22\,772--22\,789, 2022.

\bibitem{transformer}
A.~Vaswani, N.~Shazeer, N.~Parmar, J.~Uszkoreit, L.~Jones, A.~N. Gomez,
  {\L}.~Kaiser, and I.~Polosukhin, ``Attention is all you need,'' \emph{Adv.
  neural inf. proces. syst.}, vol.~30, 2017.

\bibitem{transvpr}
R.~Wang, Y.~Shen, W.~Zuo, S.~Zhou, and N.~Zheng, ``Transvpr: Transformer-based
  place recognition with multi-level attention aggregation,'' in \emph{Proc.
  IEEE Comput. Soc. Conf. Comput. Vision. Pattern. Recognit.}, pp.
  13\,648--13\,657, 2022.

\bibitem{hybrid}
Y.~Wang, Y.~Qiu, P.~Cheng, and J.~Zhang, ``Hybrid cnn-transformer features for
  visual place recognition,'' \emph{IEEE Trans. Circuits Syst. Video Technol.},
  vol.~33, no.~3, pp. 1109--1122, 2022.

\bibitem{overlaptransformer}
J.~Ma, J.~Zhang, J.~Xu, R.~Ai, W.~Gu, and X.~Chen, ``Overlaptransformer: An
  efficient and yaw-angle-invariant transformer network for lidar-based place
  recognition,'' \emph{IEEE Robot. Autom. Lett.}, vol.~7, no.~3, pp.
  6958--6965, 2022.

\bibitem{pcan}
W.~Zhang and C.~Xiao, ``Pcan: 3d attention map learning using contextual
  information for point cloud based retrieval,'' in \emph{Proc. IEEE Comput.
  Soc. Conf. Comput. Vision. Pattern. Recognit.}, pp. 12\,436--12\,445, 2019.

\bibitem{bow}
G.~Csurka, C.~Dance, L.~Fan, J.~Willamowski, and C.~Bray, ``Visual
  categorization with bags of keypoints,'' in \emph{Lect. Notes Comput. Sci.},
  vol.~1, no. 1-22, pp. 1--2.\hskip 1em plus 0.5em minus 0.4em\relax Prague,
  2004.

\bibitem{vlad}
H.~J{\'e}gou, M.~Douze, C.~Schmid, and P.~P{\'e}rez, ``Aggregating local
  descriptors into a compact image representation,'' in \emph{Proc. IEEE
  Comput. Soc. Conf. Comput. Vision. Pattern. Recognit.}, pp. 3304--3311.\hskip
  1em plus 0.5em minus 0.4em\relax IEEE, 2010.

\bibitem{cvt}
H.~Wu, B.~Xiao, N.~Codella, M.~Liu, X.~Dai, L.~Yuan, and L.~Zhang, ``Cvt:
  Introducing convolutions to vision transformers,'' in \emph{Proc. IEEE Int.
  Conf. Comput. Vision.}, pp. 22--31, 2021.

\bibitem{scancontext}
G.~Kim and A.~Kim, ``Scan context: Egocentric spatial descriptor for place
  recognition within 3d point cloud map,'' in \emph{IEEE Int. Conf. Intell.
  Rob. Syst.}, pp. 4802--4809.\hskip 1em plus 0.5em minus 0.4em\relax IEEE,
  2018.

\bibitem{pointnet}
C.~R. Qi, H.~Su, K.~Mo, and L.~J. Guibas, ``Pointnet: Deep learning on point
  sets for 3d classification and segmentation,'' in \emph{Proc. IEEE Comput.
  Soc. Conf. Comput. Vision. Pattern. Recognit.}, pp. 652--660, 2017.

\bibitem{lpdnet}
Z.~Liu, S.~Zhou, C.~Suo, P.~Yin, W.~Chen, H.~Wang, H.~Li, and Y.-H. Liu,
  ``Lpd-net: 3d point cloud learning for large-scale place recognition and
  environment analysis,'' in \emph{Proc. IEEE Comput. Soc. Conf. Comput.
  Vision. Pattern. Recognit.}, pp. 2831--2840, 2019.

\bibitem{seqot}
J.~Ma, X.~Chen, J.~Xu, and G.~Xiong, ``Seqot: A spatial--temporal transformer
  network for place recognition using sequential lidar data,'' \emph{IEEE
  Trans. Ind. Electron.}, vol.~70, no.~8, pp. 8225--8234, 2022.

\bibitem{minkloc++}
J.~Komorowski, M.~Wysocza\'nska, and T.~Trzcinski, ``Minkloc++: Lidar and
  monocular image fusion for place recognition,'' in \emph{Proc. Int. Jt. Conf.
  Neural Networks}, pp. 1--8, 2021.

\bibitem{mmdf}
X.~Yu, B.~Zhou, Z.~Chang, K.~Qian, and F.~Fang, ``Mmdf: Multi-modal deep
  feature based place recognition of mobile robots with applications on
  cross-scene navigation,'' \emph{IEEE Robot. Autom. Lett.}, vol.~7, no.~3, pp.
  6742--6749, 2022.

\bibitem{adafusion}
H.~Lai, P.~Yin, and S.~Scherer, ``Adafusion: Visual-lidar fusion with adaptive
  weights for place recognition,'' \emph{IEEE Robot. Autom. Lett.}, vol.~7,
  no.~4, pp. 12\,038--12\,045, 2022.

\bibitem{picnet}
Y.~Lu, F.~Yang, F.~Chen, and D.~Xie, ``Pic-net: Point cloud and image
  collaboration network for large-scale place recognition,'' 2020.

\bibitem{vit}
A.~Dosovitskiy, L.~Beyer, A.~Kolesnikov, D.~Weissenborn, X.~Zhai,
  T.~Unterthiner, M.~Dehghani, M.~Minderer, G.~Heigold, S.~Gelly \emph{et~al.},
  ``An image is worth 16x16 words: Transformers for image recognition at
  scale,'' \emph{arXiv preprint arXiv:2010.11929}, 2020.

\bibitem{incorporating}
K.~Yuan, S.~Guo, Z.~Liu, A.~Zhou, F.~Yu, and W.~Wu, ``Incorporating convolution
  designs into visual transformers,'' in \emph{Proc. IEEE Int. Conf. Comput.
  Vision.}, pp. 579--588, 2021.

\bibitem{pointnet++}
C.~R. Qi, L.~Yi, H.~Su, and L.~J. Guibas, ``Pointnet++: Deep hierarchical
  feature learning on point sets in a metric space,'' \emph{Adv. neural inf.
  proces. syst.}, vol.~30, 2017.

\bibitem{pct}
M.-H. Guo, J.-X. Cai, Z.-N. Liu, T.-J. Mu, R.~R. Martin, and S.-M. Hu, ``Pct:
  Point cloud transformer,'' \emph{Comput. Vis. Media.}, vol.~7, pp. 187--199,
  2021.

\bibitem{mobilenetv2}
M.~Sandler, A.~Howard, M.~Zhu, A.~Zhmoginov, and L.-C. Chen, ``Mobilenetv2:
  Inverted residuals and linear bottlenecks,'' in \emph{Proc. IEEE Comput. Soc.
  Conf. Comput. Vision. Pattern. Recognit.}, pp. 4510--4520, 2018.

\bibitem{triplet}
M.~Schultz and T.~Joachims, ``Learning a distance metric from relative
  comparisons,'' \emph{Adv. neural inf. proces. syst.}, vol.~16, 2003.

\bibitem{minkloc3d}
J.~Komorowski, ``Minkloc3d: Point cloud based large-scale place recognition,''
  in \emph{Proc. - IEEE Winter Conf. Appl. Comput. Vis., WACV.}, pp.
  1790--1799, 2021.

\end{thebibliography}

\end{document}